\documentclass[letterpaper, 10 pt, journal, twoside]{IEEEtran}

\usepackage{graphicx}
\usepackage{adjustbox}
\usepackage{amssymb}
\usepackage{amsmath}
\usepackage{mathtools}

\usepackage{amsthm}
\usepackage[acronyms, shortcuts]{glossaries}

\usepackage{wrapfig}
\usepackage{subcaption}
\usepackage{lipsum}

\usepackage{siunitx}
\usepackage{xcolor}
\usepackage{hyperref}
\usepackage[ruled,vlined,linesnumbered]{algorithm2e}
\usepackage{ifthen}
\usepackage{dsfont}
\usepackage[natbib=true,backend=biber,style=ieee,sorting=none,maxbibnames=99]{biblatex}
\usepackage{balance}
\usepackage{dirtytalk}
\usepackage{afterpage}

\usepackage{times}
\usepackage{multirow}

\usepackage[capitalize,noabbrev,nameinlink]{cleveref}
\crefformat{equation}{(#2#1#3)}
\Crefname{algocfline}{Algorithm}{Algorithms}
\Crefname{algocf}{line}{lines}

\allowdisplaybreaks

\newcommand{\U}{\mathcal{U}}

\newcommand{\X}{\mathcal{X}}

\newcommand{\numplayers}{N}

\newcommand{\norm}[1]{\left\lVert#1\right\rVert}

\newcommand{\control}{u}
\newcommand{\controls}{\mathbf{\control}}

\newcommand{\var}{x}
\newcommand{\vars}{\mathbf{\var}}

\newcommand{\consineq}{H}
\newcommand{\consineqs}{\mathbf{H}}
\newcommand{\conseq}{G}
\newcommand{\conseqs}{\mathbf{G}}

\newcommand{\lmseq}{\mathbf{w}}
\newcommand{\lmsineq}{\mathbf{v}}

\newcommand{\game}{\Gamma}

\newcommand{\tvar}{t}

\newcommand{\cost}{J}

\newcommand{\lagrangian}{\mathcal{L}}

\newcommand{\hypernormal}{\textbf{n}}
\newcommand{\hyperinter}{\textbf{m}}
\newcommand{\hyperrate}{\omega}
\newcommand{\hyperoffset}{\alpha}
\newcommand{\kozrad}{\rho}

\newcommand{\nplayers}{N}
\newcommand{\horizon}{T}

\newcommand{\tstep}{\Delta \tvar}
\newcommand{\meanmotion}{n}

\newcommand{\lmshi}{\mathbf{\lambda}_{\mathrm{hi}}}
\newcommand{\lmslo}{\mathbf{\lambda}_{\mathrm{lo}}}
\newcommand{\controlmax}{\control_{\mathrm{max}}}
\newcommand{\position}{\mathbf{p}}
\newcommand{\reconstruction}{D}
\newcommand{\mcpz}{z}

\newcommand{\mcpdimz}{n}

\newcommand{\mcp}[1]{\mathrm{MCP}(#1)}
\newcommand{\loss}{\ell}
\newcommand{\learningrate}{\alpha}

\newcommand{\mcpX}{\mathbb{X}}
\newcommand{\couples}{\Omega}
\newcommand{\couple}{\omega}
\newcommand{\weightratio}{\xi}

\newtheorem{definition}{Definition}

\newcommand{\params}{\theta}

\newacronym{admm}{ADMM}{alternating direction method of multipliers}
\newacronym{qp}{QP}{quadratic program}
\newacronym{gne}{GNE}{generalized Nash equilibrium}
\newacronym{mcp}{MCP}{mixed complementarity problem}
\newacronym{licq}{LICQ}{linear independence constraint qualification}
\newacronym{cbf}{CBF}{control barrier function}
\newacronym{ioc}{IOC}{inverse optimal control}
\newacronym{lqr}{LQR}{linear-quadratic regulator}
\newacronym{kkt}{KKT}{Karush–Kuhn–Tucker}
\newacronym{irl}{IRL}{inverse reinforcement learning}
\newacronym{mle}{MLE}{maximum likelihood estimation}
\newacronym{hji}{HJI}{Hamilton-Jacobi-Isaacs}
\newacronym[longplural=open-loop Nash equilibria,plural=OLNE]{olne}{OLNE}{open-loop Nash equilibrium}
\newacronym[longplural={partially observable Markov decision processes}]{pomdp}{POMDP}{partially observable Markov decision process}
\newacronym{svo}{SVO}{social value orientation}
\newacronym{ukf}{UKF}{unscented Kalman filter}
\newacronym{ibr}{IBR}{iterated best response}
\newacronym{awgn}{AWGN}{additive white Gaussian noise}
\newacronym{koz}{KOZ}{keep-out zone}

\bibliography{fpe-ref}

\IEEEoverridecommandlockouts

\title{\LARGE \bf Learning Hyperplanes for Multi-Robot Collision Avoidance in Space
}

\author{Fernando Palafox$^1$, Yue Yu$^2$, and David Fridovich-Keil$^1$
\thanks{Authors are with $^1$The Department of Aerospace Engineering and Engineering Mechanics and $^2$The Oden Institute for Computational Engineering \& Sciences at The University of Texas at Austin. Correspondence to \href{mailto:fernandopalafox@utexas.edu}{\tt fernandopalafox@utexas.edu}.}
\thanks{This work was supported by the National Science Foundation under Grant No. 2211548, and by the Army Research Laboratory  under Cooperative Agreement Number W911NF-23-2-0011.}
}

\begin{document}

\maketitle

\global\csname @topnum\endcsname 0
\global\csname @botnum\endcsname 0

\thispagestyle{empty}
\pagestyle{empty}

\begin{abstract}
A core challenge of multi-robot interactions is collision avoidance among robots with potentially conflicting objectives. 
We propose a game-theoretic method for collision avoidance based on rotating hyperplane constraints. 
These constraints ensure collision avoidance by defining separating hyperplanes that rotate around a keep-out zone centered on certain robots.
Since it is challenging to select the parameters that define a hyperplane without introducing infeasibilities, we propose to learn them from an expert trajectory i.e., one collected by recording human operators. 
To do so, we solve for the parameters whose corresponding equilibrium trajectory best matches the expert trajectory.
We validate our method by learning hyperplane parameters from noisy expert trajectories and demonstrate the generalizability of the learned parameters to scenarios with more robots and previously unseen initial conditions.
\end{abstract}

\section{Introduction}

Space robotics play an important role in advancing important human endeavors, such as exploration, asteroid mining, telecommunication systems, scientific research, and establishing permanent human settlements on other planets. These robots can relieve humans from undertaking tasks in remote, inhospitable, and potentially hazardous environments, such as outer space or the surfaces of celestial bodies. 

As space gets more crowded, we must manage the challenge of safe multi-robot interactions. 
On-orbit construction and/or maintenance of spacecraft, space stations, or scientific instruments will likely require close proximity interactions among many robots. 
It is crucial that we prevent collisions between these robots, particularly in situations where they may have conflicting objectives, as in construction or defense applications.

In much of the existing literature on robot collision avoidance, obstacles remain static or move in predetermined patterns that do not depend upon the robot’s decisions.
Existing methods typically bound the minimum distance between the robot and any obstacle and/or minimize a performance index that penalizes the robot for getting too close to any obstacle. 

In the case of multi-robot collision avoidance, the problem is sometimes framed as a collaborative task where all robots have the common goal of avoiding each other. 
However, these methods do not model interactions for robots with conflicting objectives. 
A concrete example of such a scenario is a complex construction task in space, where robots seek different goal positions while operating close to each other.

Multi-robot interactions can be modeled using game theory, which provides an expressive mathematical framework to describe the behavior of several robots seeking to optimize individual objectives that depend upon one another's actions. 
\emph{Dynamic} gamaes further account for strategic decision-making over time, and model the effects of sequential decisions via a dynamical system. Equilibria of these games yield state and control input trajectories that respect problem constraints and trade-off robots' individual objectives.

\begin{figure}[t!]
    \centering
    \begin{subfigure}[c]{0.49\linewidth}
        \centering
        \includegraphics[width=\linewidth]{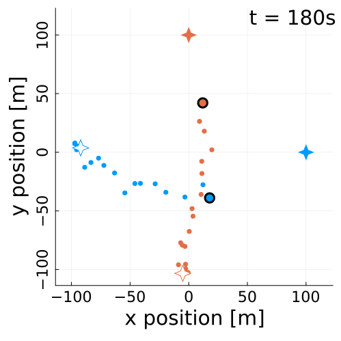}
        \caption{Noisy expert trajectory}\label{pull:subfig:expert_data}
    \end{subfigure}
    \hfill
    \begin{subfigure}[c]{0.49\linewidth}
        \centering
        \includegraphics[width=\linewidth]{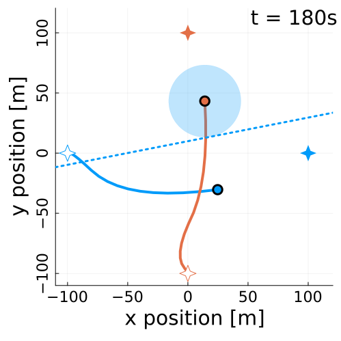}
        \caption{Learned hyperplane and KOZ}\label{pull:subfig:learned_hyper}
    \end{subfigure}
    \begin{subfigure}[c]{0.49\linewidth}
        \centering
        \includegraphics[width=\linewidth]{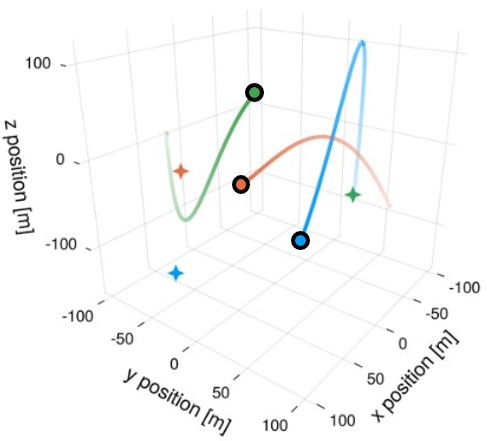}
        \caption{Collision-free trajectories}\label{pull:subfig:3d}
    \end{subfigure}
    \hfill
    \begin{subfigure}[c]{0.49\linewidth}
        \centering
        \includegraphics[width=0.5\linewidth]{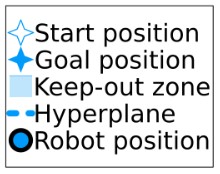}
    \end{subfigure}
    \caption{Our method uses noisy expert trajectories (\cref{pull:subfig:expert_data}) to learn the rotation rate and keep-out zone radius of a rotating hyperplane constraint (\cref{pull:subfig:learned_hyper}). We then use the same parameters to generate collision-free trajectories for a three-dimensional scenario with different initial conditions and more robots (\cref{pull:subfig:3d}).}
    \vspace{-0.5cm}
    \label{fig:pull}
\end{figure}

We propose a \emph{dynamic} game-theoretic method for multi-robot collision avoidance based on a set of rotating hyperplane constraints that define a separating hyperplane between an ego robot and an obstacle robot.
This hyperplane rotates around a circular \acrfull{koz} centered on the obstacle robot as seen in \cref{fig:Rotating_hyp} which the ego robot must avoid.
Arbitrarily selecting hyperplane parameters may introduce infeasiblities to the game such as a rotation rate that requires robots to move faster than their actuators allow, or \ac{koz} radii larger than the distance between robots at their initial positions.
Therefore, we learn the parameters from expert trajectories that already avoid collisions, which could, e.g., be collected by recording human operators.
With our method, we obtain collision-free trajectories by solving for the equilibrium trajectories of a game with learned hyperplane constraints between robots. 

We evaluate the performance of our method in a simulated collision-avoidance scenario with robot dynamics modeled by the discrete-time Hill-Clohessy-Wiltshire equations for relative orbital motion \cite{jewison2018discretecw}. 
We validate the learning capabilities of our method by learning hyperplane parameters from noisy expert data with known ground truth parameter values, and we show the generalizability of inferred parameters by applying them to different initial conditions and a scenario with more robots.
\section{Related Work}

In the domain of single-robot obstacle avoidance, \citet{DiCairano2012mpc} formulated collision-avoidance constraints in terms of half-plane constraints to facilitate safe docking maneuvers. 
Building on this, \citet{Weiss2015hyperplane} defined a rotating hyperplane, effectively separating obstacles from spacecraft trajectories. 
In this work, we extend their ideas to the multi-robot case. 

Existing game-theoretic approaches for collision avoidance include \cite{Mylvaganam2017differential} where the authors presented a framework based on differential feedback games to address multi-agent collision avoidance. Their work integrates barrier terms proportional to proximity and solves for $\epsilon$-Nash equilibria. \citet{erdogan2011gameobstacles} explored a game-theoretic strategy for formation flying and obstacle avoidance, decomposing complex avoidance tasks into smaller solvable components.
Our work is different because it ensures collision avoidance via constraints in each player's trajectory. 

In this work, we solve for unknown parameters in nonlinear constraints of an observed game-theoretic interaction (i.e., we solve the ``inverse" game). 
Relevant literature includes approaches for inverse optimal control such as \cite{menner2019constrained} where they solved for the objective and constraints of a manipulation task using Bellman's principle of optimality.  
Their work focuses on a linearly parameterized objective and selects constraints from a set of candidate constraints formed by the convex hull of all observed data points. 
Another example is \cite{menner2020mle} where they detailed two approaches for learning the parameters of an objective function based on maximum-likelihood estimation.
Solving for unknown parameters in the inequality constraints was approached in \citet{papadimitriou2023constraint} by employing an alternating optimization method that minimizes the KKT residual objective. 
Similarly, \citet{chan2020consparams} solved for the parameters of linearly parameterized inequality constraints with two duality-based approaches: the first one minimizes the duality gap and the second one seeks to satisfy strong duality. 
Note that all of these approaches focus on solving for parameters in the objective and constraints of a single expert human (or robot), whereas our work extends to the case of a multi-robot interaction with nonlinear inequality constraints. 

Existing approaches for solving an inverse game include \cite{peters2023online}, where the authors presented a method based on maximum-likelihood estimation to obtain unknown parameters in linearly parameterized objectives using noisy observations. 
Their work does not include inequality constraints but allows for collision-avoiding behavior via appropriate objective functions. 
Finally, \citet{liu2023mcp} presented a method for trajectory planning against opponents with unknown objectives and constraints by formulating the forward game as a differentiable parametric mixed complementarity problem and then using the gradient information to solve for the unknown parameters.
Their work provides the foundation for the learning algorithm we present in this paper.

\section{Problem formulation}

In this section, we describe the specifics of the multi-robot interaction, how it can be described as a game, and the Nash equilibrium solution concept.

\subsection{Robot dynamics}

We focus on the case of collision avoidance for robotic spacecraft operating in close proximity to each other. 
We model each robot's motion as a discrete-time, time-invariant dynamical system where the equations of motion are given by the discretized Hill-Clohessy-Wilshire equations for relative orbital motion \cite{jewison2018discretecw}. 
In the following definitions, an overdot denotes a time-derivative, superscripts and subscripts denote indexing by robot and time, respectively. The position of robot $i$ at time $t$ is given by $\position^i_t = \begin{bmatrix}x^i_t & y^i_t & z^i_t\end{bmatrix}^\top \in \mathbb{R}^3$ and thrust forces in each direction are given by $F_x, F_y, F_z \in \mathbb{R}$. We define the set of all possible states and controls as $\X \subset \mathbb{R}^6$ and $\U \subset \mathbb{R}^3$, respectively. Then, for robot $i$ at time $t$, state and controls are $\vars^i_t \in \X $ and $\controls^i_t \in \U $ where 
\begin{subequations}
\begin{align}
    \label{eq:state}
    &\vars_t^i :=
        \begin{bmatrix}
                    \position_t^{i^\top} & \dot{\position}_t^{i^\top}
        \end{bmatrix}^\top\\
    \label{eq:controls}
    &\controls_t^i := 
            \begin{bmatrix}
                F_{x,   t}^i & F_{y,t}^i & F_{z,t}^i
            \end{bmatrix}^\top.
\end{align}
\end{subequations}
We model discrete time as $\tvar \in [\horizon] = \{1,\dots,T\}$ and define the set of all robots as $[N] = \{1, \dots, \nplayers\}$. We use the following indexing shorthand throughout the paper: 
\begin{subequations}
    \label{shorthand}
    \begin{align*}
        \vars_t &:= 
            \begin{bmatrix}
                \vars_t^{1\top} &
                \hdots &
                \vars_t^{\nplayers\top}
            \end{bmatrix}^\top\\
        \vars &:= 
            \begin{bmatrix}
                \vars_1^{\top} & 
                \hdots & 
                \vars_\horizon^{\top}
            \end{bmatrix}^\top\\
        \vars^{i} &:= 
            \begin{bmatrix}
                \vars_{1}^{i\top} &
                \hdots &
                \vars_{\horizon}^{i\top}
            \end{bmatrix}^\top\\
        \vars^{-i} &:= 
            \begin{bmatrix}
                \vars^{1\top} &
                \hdots & 
                \vars^{(i - 1)\top}&
                \vars^{(i + 1)\top}&
                \hdots&
                \vars^{N\top}
            \end{bmatrix}^\top
    \end{align*}
\end{subequations}

We describe state dynamics with the time-invariant difference equation 
\begin{equation}
\label{eq:dynamics}
    \vars^i_{t+1} = A\vars^i_\tvar + B\controls_\tvar^i,
\end{equation}
where $A \in \mathbb{R}^{6\times6}$ and $B \in \mathbb{R}^{3\times3}$ are given in \cref{appendix:eom}. By defining 
$\conseq^i_t := \vars^i_{t+1} - (A\vars^i_\tvar + B\controls_\tvar^i)$, the equality constraints encoding dynamic feasibility for robot $i$ are
\begin{equation}
\conseq^i := 
\begin{bmatrix}
    \conseq^{i\top}_1 &
    \hdots &
    \conseq^{i\top}_{\horizon - 1} 
\end{bmatrix}^\top = \mathbf{0}.
\end{equation}
Note that $\conseq^i$ is implicitly parameterized by the initial state $\vars_1$.

We model finite thrusting capabilities by limiting thrust magnitude to $\controlmax \in \mathbb{R}^+$ in any direction. Mathematically, this is given by the constraints
\begin{subequations}
    \begin{align}
        \controlmax \mathbf{1}_3 +\controls &\geq 0\\
        \controlmax \mathbf{1}_3 -\controls &\geq 0
    \end{align}
\end{subequations}
where $\mathbf{1}_{3}$ denotes a $3\times1$ vector of ones, and $\geq$ is evaluated element-wise. 

\subsection{Objective}
The objective of every robot is to minimize the distance between its position at final time $T$ and its goal position while also minimizing control effort. Mathematically, we define it as a scalar function $\cost^i: \mathbb{R}^{6\numplayers\horizon} \times \mathbb{R}^{3\numplayers\horizon} \to \mathbb{R}$, 
\begin{equation}
    \cost^i(\vars, \controls)
    := \norm{
            \position_T^i - \position_\mathrm{goal}^i
          }_2^2
    + \weightratio^i\norm{\controls^i}_2^2,
\end{equation}
where $\position_\mathrm{goal}^i \in \mathbb{R}^3$ is robot $i$'s goal position, and $\weightratio^i \in \mathbb{R^+}$ scales the weight of the second term relative to the weight of the first term.
\begin{figure}[h!]
    \centering
    \includegraphics[width=0.8\linewidth]{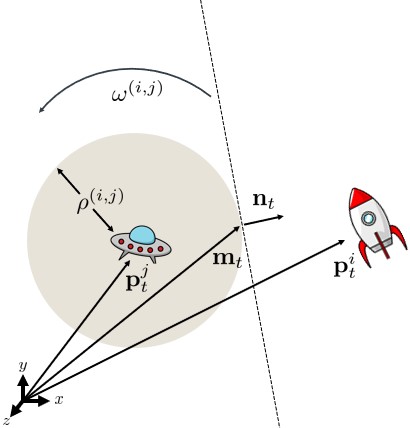}
    \caption{For constraint $\consineq^{(i,j)}$, the hyperplane (dotted line) rotates at a rate $\omega^{(i,j)}$ around the shaded \ac{koz} with radius $\kozrad^{(i,j)}$.}
    \label{fig:Rotating_hyp}
    \vspace{-0.5cm}
\end{figure}

\subsection{Rotating hyperplane constraints}
In the interest of computational efficiency, we follow \cite{Weiss2015hyperplane} and formulate collision avoidance in terms of an affine hyperplane constraint that changes with time. 
We define $\couples$ as the set of robot pairs for which a hyperplane constraint exists in the game. 
For example, $\couples = \{(1,2), (1,3), (2,3)\}$ in a game with hyperplanes separating robots $1$ and $2$, $1$ and $3$, and $2$ and $3$.
We denote $\couples^i$ as the set of pairs that include robot $i$. For example, in the previous game $\couples^2 = \{(1,2), (2,3)\}$. 
Robot pair $(i,j)$ avoids collisions by defining a separating hyperplane that rotates at an angular velocity $\hyperrate^{(i,j)} \in \mathbb{R}$ around a circular \ac{koz} of radius $\kozrad^{(i,j)} \in \mathbb{R}^+$ centered on robot $j$, as illustrated in \cref{fig:Rotating_hyp}. 
Mathematically, the constraint is defined as
\begin{subequations}
    \begin{align}
        &\consineq_t^{(i,j)} \geq 0, \forall \tvar \in \{2, \hdots, \horizon\}\label{eq:single_hyp_constraint}\\
        &\mathrm{where~}\consineq_t^{(i,j)} := \hypernormal_t^\top(\position_t^{i} - \hyperinter_t) \\
        &\qquad\quad\hypernormal_t := 
            \begin{bmatrix}
                \cos(\hyperoffset^{(i,j)} + \omega^{(i,j)} (t-1))\\
                \sin(\hyperoffset^{(i,j)} + \omega^{(i,j)} (t-1))
            \end{bmatrix}\\
        &\qquad\quad\hyperinter_t := \position_t^i + \kozrad^{(i,j)} \hypernormal_t.
    \end{align}
\end{subequations}
$\hyperoffset^{(i,j)} \in \mathbb{R}$ is the angle between the $x$-axis and $\position_t^i - \position_t^j$. 
Note that this formulation only depends on the planar components of the state. 
This limits the range of possible trajectories since robots' planar positions cannot overlap even if out-of-plane positions are at a safe distance. 
This trade-off can be justified by the reduced complexity of the constraint formulation and by noting that, in practice, this formulation is sufficient for collision-free trajectories in common space maneuvers e.g., docking \cite{DiCairano2012mpc}.

In general, this interaction is non-cooperative because robots' objectives are arbitrary and can conflict. However, we assume that all agents are responsible for avoiding collisions, and therefore, robot $i$'s complete set of constraints contains constraints for robots it wishes to avoid (e.g. $\consineq^{(i,j)}$, a hyperplane centered on robot $j$ that robot $i$ wishes to avoid), as well as any constraints for robots trying to avoid robot $i$ itself (e.g. $\consineq^{(k,i)}$, a hyperplane centered on robot $i$ that robot $k$ wishes to avoid). 
Therefore, robot $i$'s hyperplane constraints are expressed as
\begin{subequations}
\begin{align}
    \label{eq:cons_inequality}
    \consineq^{(i,j)} &:=  
        \begin{bmatrix}
            \consineq_2^{(i,j)} & \hdots & \consineq_\horizon^{(i,j)}
        \end{bmatrix}^\top\\
    \consineqs^{i} &:= \{\consineq^{\couple} \mid  \couple \in \couples^i\} \geq 0
    \label{eq:set_builder}
\end{align}
\end{subequations}
where the set-builder notation in \cref{eq:set_builder} is 
abused to return a vector $\consineqs^i \in \mathbb{R}^{(\horizon - 1)|\couples|}$ and $\geq$ is evaluated element-wise. 

\subsection{Collision avoidance game}
We model the multi-robot interaction with hyperplane constraints and thrust limits as an open-loop and infinite game $\game(\params) := (\vars_1,\{\cost^i\},\{\conseqs^i\},\{\consineqs^i\},\controlmax,\params,[\nplayers], [\horizon])$, where \(\params := \begin{Bmatrix}
                        \hyperrate,& 
                        \kozrad,& 
                        \weightratio
                    \end{Bmatrix}\) contains the hyperplane parameters and cost weights for all robots. Each robot is only given the initial conditions $\vars_1$ and then solves the coupled optimization problems
\begin{subequations}
\label{eq:game}
\begin{align}
    \label{eq:game_objective}
    &\vars^{i*},\controls^{i*} \in \arg \min_{\vars^i,\controls^i} \cost^{i}(\vars,\controls)\\
    &\mathrm{subject~to}~ \conseqs^i = \mathbf{0}\\
    &\qquad \qquad ~~ \consineqs^i \geq \mathbf{0}\label{eq:game_constraints_inequality}\\
    &\qquad \qquad ~~\control_{\mathrm{max}}\mathbf{1}_3 +\controls \geq 0\label{eq:game_constraints_thrust_lo}\\ 
    &\qquad \qquad ~~\control_{\mathrm{max}}\mathbf{1}_3 -\controls \geq 0\label{eq:game_constraints_thrust_hi}.
\end{align}
\end{subequations} 
Note that this problem is implicitly parameterized by $\params$, which enters both $\cost^i$ and $\consineqs^i$.


We seek \ac{gne} solutions \cite{Facchinei2004variation} in which no robot has a unilateral incentive to change its current strategy. 
This concept captures the non-cooperative nature of the interaction between rational robots with potentially conflicting objectives and is defined as follows: 

\begin{definition}[Generalized Nash equilibrium]
A set of strategies $\controls^* := \{\controls^{1*}, \hdots, \controls^{N*}\}$ is a generalized Nash equilibrium of the game $\game(\params)$ if it satisfies
\begin{equation}
\label{eq:ne_conditions}
    \cost^i(\vars^*, \controls^*) \leq \cost^i(\vars,\{\controls^i, \controls^{-i*}\}), \forall i \in [\nplayers]
\end{equation}
for any feasible deviation $(\vars, \controls)$ with $\controls^{-i*}$ denoting all but robot $i$'s controls.
\end{definition}

In general, computing Nash equilibria is computationally intractable \cite{daskalakis2009nashcomplexity}, so we solve for trajectories $(\vars, \controls)$ that locally satisfy the equilibrium conditions $\cref{eq:ne_conditions}$ to first order. Assuming linear independence of the constraints, solutions for \cref{eq:game} must satisfy the intersection of every player's \ac{kkt} conditions \cite{Facchinei2004variation}. To express these, we introduce Lagrange multipliers $\lmseq, \lmsineq, \lmshi, \lmslo$ and write robot $i$'s Lagrangian as 
\begin{equation}
\begin{split}
\mathcal{L}^i = ~
 &\cost^i - \lmseq^{i\top}\conseqs^i - \lmsineq^{i\top}\consineqs^i\\ 
 &- \lmshi^{i\top}(\control_{\mathrm{max}}\mathbf{1}_3 +\controls^i) - \lmslo^{i\top}(\control_{\mathrm{max}}\mathbf{1}_3 -\controls^i)
\end{split}
\end{equation}
where $\lmsineq^{i} = \{\lmsineq^{\couple} \mid  \couple \in \couples^i\}$.
Then, the \ac{kkt} conditions are
\begin{equation}
\label{eq:kkt}
    \forall i \in [\nplayers]
    \left\{
    \begin{array}{l}
        \nabla_{\vars^i}\lagrangian^i = 0\\
        \nabla_{\controls^i}\lagrangian^i = 0\\ 
        \conseqs^i = 0\\
        0 \leq \consineqs^i \bot \lmsineq^i \geq 0\\
        0 \leq \control_{\mathrm{max}}\mathbf{1}_3 +\controls ~ \bot ~ \lmshi \geq 0\\
        0 \leq \control_{\mathrm{max}}\mathbf{1}_3 -\controls ~ \bot ~ \lmslo \geq 0.\\
    \end{array}
    \right.
\end{equation}
Robots share multipliers associated with hyperplane constraints $\consineqs^i$ to encode shared responsibility for constraint feasibility. 
For example, $(i,j) \in \couples$ implies that $\consineq^{(i,j)} \in \consineqs^i \cap \consineqs^j$ so  $\lmsineq^i$ and $\lmsineq^j$ will share some elements.

It can be shown that solving \cref{eq:kkt} is equivalent to solving a \ac{mcp} \cite{Facchinei2004variation}, parameterized by $\params$ and denoted as $\mcp{\params}$. 
The solution for $\mcp{\params}$ is a vector $\mcpz^* \in \mathbb{R}^\mcpdimz$ that can be expressed in terms of the decision variables in \cref{eq:kkt} as 
\begin{equation}
    \mcpz^{*} = [\vars^{*\top}, \controls^{*\top}, \lmseq^{*\top}, \lmsineq^{*\top}, \lmshi^{*\top}, \lmslo^{*\top}]^\top. 
\end{equation}
This equivalence is very convenient because, as shown in \cite{liu2023mcp}, it is possible to differentiate $\mcpz^*$ with respect to $\params$. 
This fact will be fundamental in the development of the hyperplane parameter learning algorithm, as discussed in the next section. 

\section{Learning Hyperplane Parameters}
Selecting hyperplane parameters without introducing infeasibilities into the game can be challenging, particularly due to the thrust limit constraints \labelcref{eq:game_constraints_thrust_lo,eq:game_constraints_thrust_hi}.
In games with two robots, selecting parameters can be done through judicious trial and error.
However, the increasingly complicated geometry rules out this approach for games with more robots.
Instead, we propose to learn these parameters from expert trajectories known to satisfy all constraints, such as the trajectory of two small satellites controlled by expert human operators. 

Concretely, we begin with an observed state trajectory $\tilde{\vars}$, assumed to represent the equilibrium behavior of a parametric collision-avoidance game $\game(\params)$ (where all or only some of the elements of $\params$ are unknown), and then solve an ``inverse'' game to recover the unknown parameters. 
The solution to the inverse game is the set of constraint parameters for which the resulting equilibrium best matches the observed trajectories, i.e.
\begin{equation}
    \label{eq:inference_problem}
    \min_{\params, \vars} \loss(\vars,\tilde{\vars}) \mathrm{~subject~to~} \vars \in \mcpX 
\end{equation}
where $\loss: \mathbb{R}^{6\numplayers\horizon} \times \mathbb{R}^{6\numplayers\horizon} \to \mathbb{R}, \loss(\vars,\tilde{\vars}) := \norm{\vars - \tilde{\vars}}^2_2$, and 
\begin{equation}
    \begin{split}
        \mcpX = \{\vars \in \mathbb{R}^{6\numplayers\horizon} \mid \exists \controls, \lmseq, \lmsineq, \lmshi, \lmslo, \mathrm{~such~that~} \\
        \mcpz = 
    \begin{bsmallmatrix}
        \vars^{\top},  \controls^{\top}, \lmseq^{\top}, \lmsineq^{\top}, \lmshi^{\top}, \lmslo^{\top}
    \end{bsmallmatrix}^\top
        \mathrm{~solves~}\mcp{\params}\}
    \end{split}
\end{equation}

We follow the approach in \cite{liu2023mcp} and solve for the hyperplane parameters by descending the gradient of $\loss$ with respect to $\params$ as shown in \cref{alg:learn_hyperplane}. We found that scaling the gradient with an Adam optimizer \cite{Kingma2017adam} greatly improved numerical stability. 

\begin{algorithm}[ht]
\SetAlgoLined
\DontPrintSemicolon
\caption{Learn hyperplane parameters}\label{alg:learn_hyperplane}
{
\small
    \textbf{Input:} Initial $\theta$, learning rate $\learningrate$, decay rates $\beta_1, \beta_2$, convergence tolerance $\epsilon>0$.
    
    \While{$\norm{\nabla_\params \loss} \geq \epsilon$}{
        $(\mcpz^*, \nabla_\params\mcpz^*) \leftarrow $
        $\text{solve}~\mcp{\params}$\\
        $\nabla_\params\loss \leftarrow \nabla_\vars\loss^\top \nabla_{\mcpz^*}\vars \nabla_\theta\mcpz^* $\\
        $\params \leftarrow \params - \text{Adam}(\nabla_\params \loss; \alpha, \beta_1, \beta_2)$\\
    }

    \Return{$\params$}
}
\vspace{-0.25em}
\end{algorithm}

In the case of rotating hyperplane constraints, gradient descent yields a loss-optimal set of hyperplane parameters without requiring that all hyperplane constraints be active.
If a constraint is never active, the gradient of the loss with respect to the constraint parameters will be zero (even if the parameters are not optimal), rendering gradient descent useless. 
We avoid this issue because a set of hyperplane parameters affects $\horizon - 1$ constraints as defined in \cref{eq:single_hyp_constraint}. 
Although most of these constraints are not active, at least one of them must be if a collision is avoided because of the hyperplane constraints.
Therefore, the gradient of the loss function with respect to the hyperplane parameters will only be zero at optimal points or in the case where the hyperplane constraint is never active at any point in the game. 
In practice, the latter case was very rare and never occurred in the Monte Carlo experiments presented in \cref{sec:experiments}.
Therefore, we can use gradient descent to converge on a set of loss-optimal parameters.

We solve \cref{eq:inference_problem} with a Julia implementation of \cref{alg:learn_hyperplane} that is publicly available.\footnote{\href{https://github.com/CLeARoboticsLab/InverseHyperplanes.jl}{github.com/CLeARoboticsLab/InverseHyperplanes.jl}}
We express the \ac{mcp} using ParametricMCPs.jl \cite{Peters2023parametric}, a mathematical programming layer that compiles a differentiable \ac{mcp} parameterized by a runtime vector. 
Within ParametricMCPs.jl, the \ac{mcp} is solved via the PATH solver \cite{dirkse1995path}.
\section{Experimental Results}
\label{sec:experiments}

In this section, we present experimental results to demonstrate the performance of our method. 
First, we validate learning capabilities by presenting the performance of our method when learning from noise-corrupted expert trajectories.  
Then, we demonstrate the generalizability of the learned parameters by presenting a six-robot collision-free trajectory using the parameters learned from a two-robot expert trajectory. 
Finally, we characterize the sensitivity of the learned parameters to perturbations in the initial velocity of the robots. 

\subsection{Learning from noisy expert trajectories}

We validate the performance of \cref{alg:learn_hyperplane} by presenting the difference between ground truth hyperplane parameters and parameters learned from a noisy expert trajectory. 
For the expert trajectory in this experiment, we use an equilibrium state trajectory for a two-dimensional, two-robot game with initial conditions and ground truth parameters given in \cref{table:parameters} of \cref{appendix:parameters}. 
Two-dimensional dynamics are trivially implemented by noting that the out-of-plane position dynamics in  \cref{eq:dynamics} are decoupled from the planar position.

\begin{figure}[t!]
    \centering
    \begin{subfigure}[b]{0.49\linewidth}
        \centering
        \includegraphics[width=\linewidth]{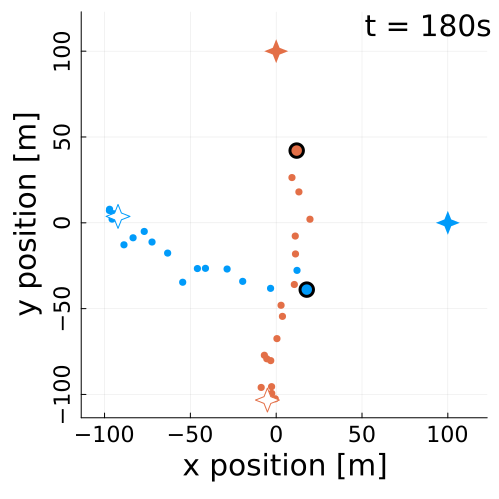}
        \caption{Noisy trajectory ($\sigma = 5$)}\label{subfig:expert_data}
    \end{subfigure}
    \hfill
    \begin{subfigure}[b]{0.49\linewidth}
        \centering
        \includegraphics[width=\linewidth]{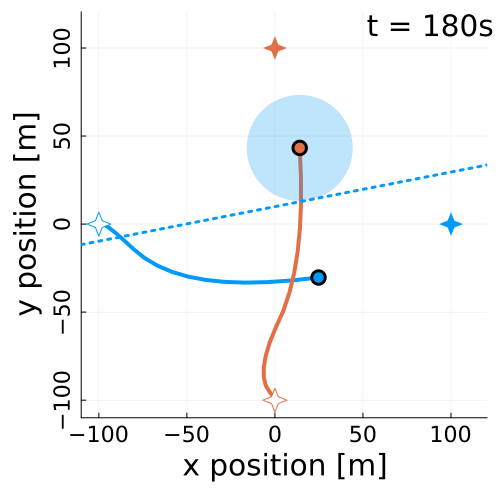}
        \caption{Learned hyperplane}\label{subfig:expert_hyperplane}
    \end{subfigure}
    \caption{Using noisy expert trajectories, our method learns hyperplane parameters and recovers the corresponding equilibrium trajectory. Legend in \cref{fig:pull}.}
    \vspace{-0.25cm}
    \label{fig:expert_data_hyperplane}
\end{figure}

We generate $400$ noise-corrupted trajectories by adding isotropic additive white Gaussian noise to the expert state trajectory at 20 levels of standard deviation $\sigma$ from $0.0$ to $20.0$ and $20$ samples for each level. 
\cref{subfig:expert_data} shows a snapshot of a typical noisy state trajectory, and \cref{subfig:expert_hyperplane} shows the learned hyperplane and corresponding equilibrium trajectory.

\begin{figure}[h!]
    \centering
    \includegraphics[width=\linewidth]{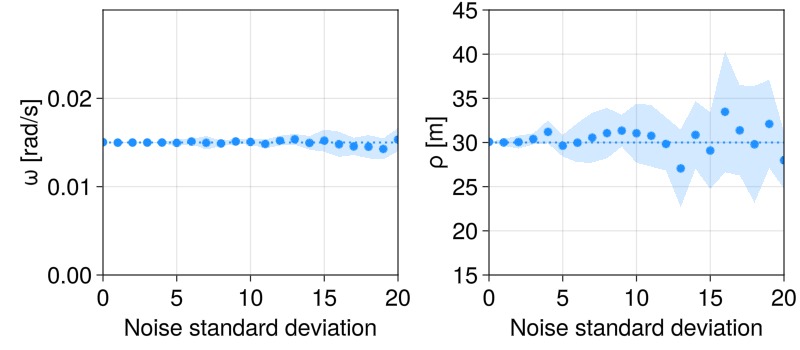}
    \caption{Learned hyperplane parameters vs. noise level. A dotted line is the ground truth value, dots are median inferred values, and ribbons show the interquartile range.}
    \label{fig:bands}
    \vspace{-0.25cm}
\end{figure}

We show the accuracy of the learned hyperplane parameters by presenting a comparison between ground-truth and learned parameters in \cref{fig:bands} as a function of noise level.
Learned rotation rates are very close to the ground truth even when the trajectory is corrupted by extreme noise levels. 
The learned \ac{koz} radius is not as accurate as noise increases, but this is likely because the solution is not as sensitive to changes in radius $\kozrad$ as it is to changes in rotation rate $\hyperrate$.
That is, changes in $\hyperrate$ require a larger change in the trajectory than changes in $\kozrad$, and large changes to the controls may be infeasible due to actuator limits (\cref{eq:game_constraints_thrust_lo,eq:game_constraints_thrust_hi}).


\begin{figure}[h!]
    \centering
    \includegraphics[width=\linewidth]{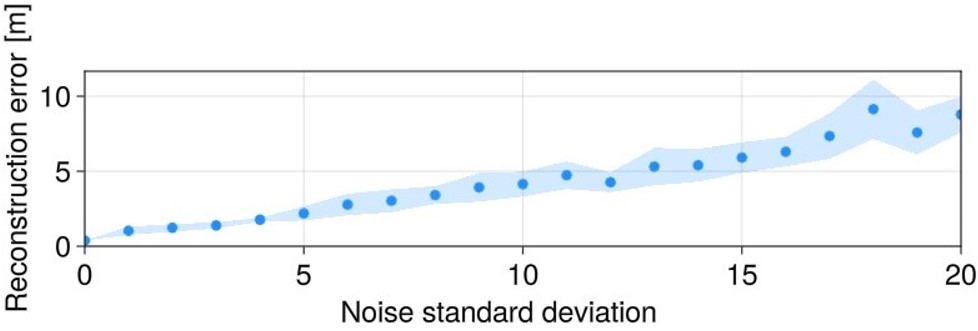}
    \caption{Reconstruction error \cref{eq:reconstruction} vs. noise level. Dots are median error values, and ribbons show the interquartile range.}
    \label{fig:reconstruction}
\end{figure}

We show that the learned parameters \cref{eq:inference_problem} correspond to an equilibrium trajectory that closely matches ground truth by computing a measure of its difference from the noise-free expert trajectory. 
Concretely, we report the reconstruction error $\reconstruction$:
\begin{equation}
    \reconstruction(\params_\mathrm{truth}, \params_\mathrm{learned}) = \frac{1}{\nplayers \horizon}\sum_{i \in [\nplayers]}\sum_{\tvar \in [\horizon]}\norm{
    \position_{\mathrm{truth},t}^i - \position_{\mathrm{learned},t}^i}^2_2
    \label{eq:reconstruction}
\end{equation}
where $\position_\mathrm{{\mathrm{truth},t}}^i$ is robot $i$'s position at time $\tvar$ as given by the noise-free expert data with ground truth parameters $\params_\mathrm{truth}$, and  $\position_{\mathrm{learned},\tvar}^i$ is robot $i$'s position at time $\tvar$ as given by a Nash equilibrium of a game parameterized by the learned parameters $\params_\mathrm{learned}$. 
\cref{fig:reconstruction} shows the median value of $\reconstruction$ for all trials as well the associated interquartile range.
As expected, the error increases as noise increases. 
However, at reasonable noise values ($\sigma<5$), reconstruction error is relatively low, particularly when considering the scale of the problem ($200\unit{\metre}$ between initial and goal positions). 

\subsection{Generalizability of learned parameters}

\begin{figure}[h!]
    \centering
    \includegraphics[width=0.5\linewidth]{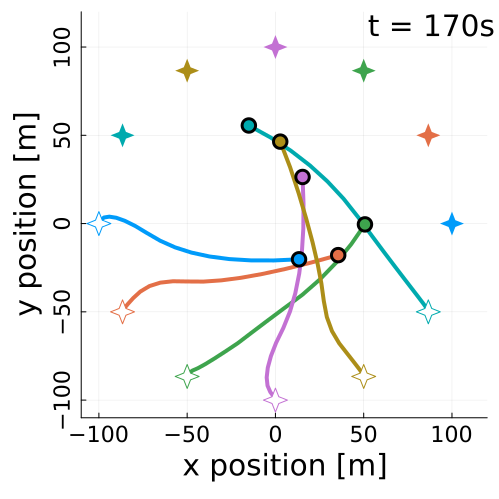}
    \caption{Collision-free trajectory for a six-robot game. Hyperplanes and \ac{koz}s are not shown for clarity. We adjust initial and goal positions, add additional pairs to $\couples$, and otherwise reuse the parameters in \cref{appendix:parameters}.}
    \label{fig:forward_6p}
    \vspace{-0.25cm}
\end{figure}

Hyperplane parameters from expert trajectories generalize to scenarios with more robots. 
For example, one can use the parameters learned from a two-robot expert trajectory to generate a collision-free trajectory for a six-robot game, as shown in \cref{fig:forward_6p}. 
Our method is easily extended to games with more robots by simply appending the KKT conditions for each additional robot to $\cref{eq:kkt}$ and defining an appropriate set of robot pairs $\couples$.

\begin{figure}[h!]
    \centering
    \includegraphics[width=\linewidth]{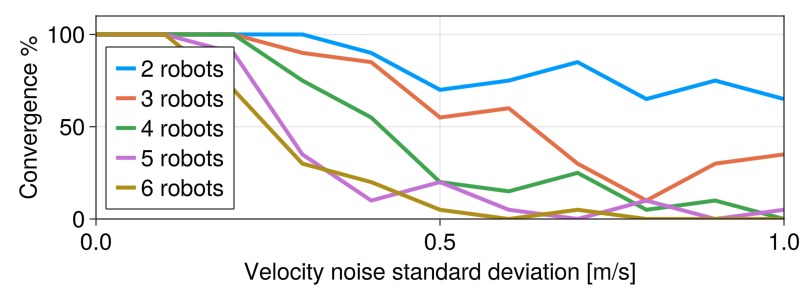}
    \caption{}
    \label{fig:initial_convergence}
    \caption{Success of collision-free trajectory generation as a function of velocity noise level. The initial velocity of the expert trajectory is zero and we ran $20$ trials at each noise level.}
    \vspace{-0.25cm}
\end{figure}

Learned parameters can also be used to generate collision-free trajectories for scenarios with previously unseen initial conditions. 
Anecdotally, trajectory generation from learned parameters is robust to changes in initial position but sensitive to changes in initial velocities.  
To quantify this sensitivity, we perturb the initial velocity with noise at varying levels and then attempt to construct a collision-free trajectory.  
In \cref{fig:initial_convergence} we report trajectory generation convergence success as a function of velocity noise level given a nominal velocity. 
In this experiment, the expert trajectory has an initial velocity of zero.

Results show that learned parameters readily generalize to scenarios with more robots as long as the initial velocity is close to the initial velocity of the expert trajectories. 
This may not be an issue under the assumption that conditions encountered in practice will be similar to those in the expert trajectories. 
We also note that the parameters are more sensitive to changes in initial velocity as we include more robots. 
This is likely because the more robots we include, the more crowded the space becomes, and it naturally becomes harder to optimize their trajectory. 

Finally, since the hyperplanes only constrain planar positions and the dynamics of the out-of-plane positions are decoupled from the planar dynamics, collision-free trajectory generation can be trivially extended to three dimensions.
We present an example trajectory in \cref{pull:subfig:3d}.

\section{Conclusion}
We propose a game-theoretic method for multi-robot collision avoidance in space robotic applications.
Our method is different from existing techniques as it extends collision avoidance with rotating hyperplanes to a multi-robot setting, using parameters learned from expert trajectories. 
Unlike previous inverse game solvers, we explicitly solve for parameters in the inequality constraints.

We conduct extensive numerical simulations to validate learning capabilities from noise-corrupted expert trajectories and demonstrate the generalizability of the learned parameters to games with different initial conditions and/or more robots.
Future areas of work include applying this method to other dynamical systems and using real expert trajectories like tracked on-orbit maneuvers. 

\printbibliography

\crefalias{section}{appendix}
\begin{appendices}
\section{Robot dynamics}
\label{appendix:eom}
Here we present the complete robot dynamics \cref{eq:dynamics} as a linear-time invariant system of equations representing the discrete-time Hill-Clohessy-Wiltshire equations for relative orbital motion as defined in \citet{jewison2018discretecw}. 
This system describes the robot's state as a function of applied controls and state at the previous time step.
\begin{subequations}
    \begin{align*}
    &\vars^i_{t+1} = A\vars^i_\tvar + B\controls_\tvar^i\\
    A &= 
        \left[\begin{smallmatrix}
            4 - 3\cos(\meanmotion\tstep) & 0 & 0 \\ 
            6(\sin(\meanmotion\tstep) - \meanmotion\tstep) & 1 & 0 \\
            0 & 0 & \cos(\meanmotion\tstep) \\
            3\meanmotion\sin(\meanmotion\tstep) & 0 & 0 \\ 
            -6\meanmotion(1 - \cos(\meanmotion\tstep)) & 0 & 0 \\
            0 & 0 & -\meanmotion\sin(\meanmotion\tstep) \\
        \end{smallmatrix}\right. \ \cdots\\
        &\; \cdots \ \left.\begin{smallmatrix}
            \frac{1}{\meanmotion}\sin(1 - \meanmotion\tstep) & \frac{2}{\meanmotion}(1 - \cos(\meanmotion\tstep)) & 0 \\
            -\frac{2}{\meanmotion}(1 - \cos(\meanmotion\tstep)) & \frac{1}{\meanmotion}(4\sin(\meanmotion\tstep) - 3\meanmotion\tstep) & 0\\
             0 & 0 & \frac{1}{\meanmotion}\sin(\meanmotion\tstep)\\
             \cos(\meanmotion\tstep) & 2\sin(\meanmotion\tstep) & 0\\
             -2\sin(\meanmotion\tstep) & 4\cos(\meanmotion\tstep) - 3 & 0\\
             0 & 0 & \cos(\meanmotion\tstep)
        \end{smallmatrix} \right]\\
        &\\
    B &= \frac{1}{m}
        \left[\begin{smallmatrix}
             \frac{1}{\meanmotion}\sin(\meanmotion\tstep)                          & \frac{2}{\meanmotion}(1 -\cos(\meanmotion\tstep))                          \\
            -\frac{2}{\meanmotion^2}(\meanmotion\tstep - \sin(\meanmotion\tstep)) & \frac{4}{\meanmotion^2}(1 - \cos(\meanmotion\tstep)) - \frac{3}{2}\tstep^2 \\
            0                                                                     & 0 \\
            \frac{1}{\meanmotion}\sin(\meanmotion\tstep) & \frac{2}{\meanmotion}(1 - \cos(\meanmotion\tstep) \\
            -\frac{2}{\meanmotion}(1 - \cos(\meanmotion\tstep)) & \frac{4}{\meanmotion}\sin(\meanmotion\tstep) - 3\tstep \\
            0 & 0 \\
        \end{smallmatrix}\right. \ \cdots\\
        &\qquad \qquad \qquad \qquad \qquad \qquad \;\, \cdots \left. 
        \begin{smallmatrix}
            0 \\
            0 \\
            \frac{1}{\meanmotion^2}(1 - \cos(\meanmotion\tstep)) \\
            0 \\ 
            0 \\
            \frac{1}{\meanmotion}\sin(\meanmotion\tstep)
        \end{smallmatrix} \right]
    \end{align*}
\end{subequations}
where $\tstep$ is the discretization interval and $m$ is the satellite mass. $\meanmotion \approx \sqrt{G/a}$ is the angular speed required for a body to complete one orbit (mean motion) and is defined in terms of the universal constant of gravitation $G$ and the orbit semi-major axis $a$. We assume a circular orbit around Earth, in this case, $a = (\mathrm{orbital~altitude} + \mathrm{~Earth's~radius})$.
\vfill\eject

\section{Simulation Parameters}
In \cref{table:parameters}, we include relevant simulation parameters for the two-dimensional, two-robot inverse game solved in \cref{sec:experiments}. 
In the six-robot trajectory in \cref{fig:forward_6p}, we adjust initial and goal positions, add additional pairs to $\couples$, and reuse parameters otherwise.
\label{appendix:parameters}

\begin{table}[ht!]

    \begin{center}
    {\renewcommand{\arraystretch}{1.3}
    \begin{tabular}{ |p{1.0cm}||c|c|c| } 
    \hline
      & \textbf{Parameter} & \textbf{Value} & \textbf{Units} \\
    \hline \hline
    \multirow{9}{5em}{\textbf{Game}}  & Players & 2 & \\ 
                                        & Initial state & $[\begin{smallmatrix}0&100&0&0&-100&0&0&0\end{smallmatrix}]^\top$ & m \\ 
                                        & Goal positions & $[\begin{smallmatrix}0&-100\end{smallmatrix}]^\top, [\begin{smallmatrix}100&0\end{smallmatrix}]^\top$& m \\
                                        & Orbital altitude & 400 & km\\ 
                                        & Satellite mass & 100 & kg\\ 
                                        & $\control_{\mathrm{max}}$ & 1.0 & N \\
                                        & $\weightratio^i$ & 0.0001 & \\ 
                                        & $\couples$ & $\{(1,2)\}$ & \\ 
                                        & $\tstep$ & 5.0 & s\\ 
                                        & Total time & 220 & s\\
    \hline
    \multirow{2}{5em}{\textbf{Ground truth}} & $\hyperrate^{(1,2)}$   & 0.015 & rad/s \\  
                                           & $\kozrad^{(1,2)}$      & 30.0 & m\\ 
    
    \hline
    \multirow{3}{5em}{\textbf{Inverse game}} & Initial guess $\hyperrate^{(1,2)}$ & $0.008$ & rad/s\\ 
                                        & Initial guess $\kozrad^{(1,2)}$ & $10.0$ & m\\ 
                                        & Num. of grad. steps & 30 & \\
    \hline
    \multirow{2}{5em}{\textbf{Experi- ments}} & $\sigma$ range & $0:5:40$ & m \\  
                                           & Trials & 20 & \\ 
    \hline
    \end{tabular}}
    \caption{Simulation parameters for hyperplane learning using noisy expert trajectories. The initial guess for the hyperplane rotation rate has empirically been shown to work well in a single-robot scenario \cite{Zagaris2018hyperparams}.}
    \label[table]{table:parameters}
    \end{center}
    \vspace{-0.5cm}
\end{table}
\end{appendices}




\end{document}